\documentclass[letterpaper, 10 pt, journal, twoside]{IEEEtran}

\usepackage{graphicx, tipa}
\usepackage{amsmath,amssymb,amsfonts}
\usepackage[caption=false]{subfig}
\usepackage[ruled,linesnumbered, noend]{algorithm2e}
\usepackage{dirtytalk}
\usepackage{hyperref}
\usepackage{gensymb}
\usepackage{tabularx}
\usepackage{multirow}
\usepackage[short]{optidef}
\usepackage{booktabs}
\usepackage{siunitx}
\usepackage{titlesec}
\newcounter{cases}
\newcounter{subcases}[cases]

\makeatletter
\newcommand{\removelatexerror}{\let\@latex@error\@gobble}
\makeatother
\SetKwComment{Comment}{$\triangleright$\ }{}

\newcommand\Tstrut{\rule{0pt}{2.0ex}}         

\title{ 
Onboard dynamic-object detection and tracking for autonomous robot navigation with RGB-D camera}

\author{Zhefan Xu\footnotemark*, Xiaoyang Zhan\footnotemark*, Yumeng Xiu, Christopher Suzuki, and Kenji Shimada 
\thanks{*The authors contributed equally.}
\thanks{Manuscript received: July 4 2023; Revised: October 1 2023; Accepted: November 11 2023.}
\thanks{This paper was recommended for publication by Editor Pascal Vasseur upon evaluation of the Associate Editor and Reviewers' comments.}
\thanks{Zhefan Xu, Xiaoyang Zhan, Yumeng Xiu, Christopher Suzuki, and Kenji Shimada are with the Department of Mechanical Engineering, Carnegie Mellon University, 5000 Forbes Ave, Pittsburgh, PA, 15213, USA. {\tt\footnotesize zhefanx@andrew.cmu.edu}}
\thanks{Digital Object Identifier (DOI): see top of this page.}
}

\begin{document}
\markboth{IEEE Robotics and Automation Letters. Preprint Version. Accepted November, 2023}
{Xu \MakeLowercase{\textit{et al.}}: Onboard dynamic-object detection and tracking for autonomous robot navigation with RGB-D camera} 

\maketitle

\begin{abstract}
Deploying autonomous robots in crowded indoor environments usually requires them to have accurate dynamic obstacle perception. Although plenty of previous works in the autonomous driving field have investigated the 3D object detection problem, the usage of dense point clouds from a heavy Light Detection and Ranging (LiDAR) sensor and their high computation cost for learning-based data processing make those methods not applicable to small robots, such as vision-based UAVs with small onboard computers. To address this issue, we propose a lightweight 3D dynamic obstacle detection and tracking (DODT) method based on an RGB-D camera, which is designed for low-power robots with limited computing power. Our method adopts a novel ensemble detection strategy, combining multiple computationally efficient but low-accuracy detectors to achieve real-time high-accuracy obstacle detection. Besides, we introduce a new feature-based data association and tracking method to prevent mismatches utilizing point clouds' statistical features. In addition, our system includes an optional and auxiliary learning-based module to enhance the obstacle detection range and dynamic obstacle identification. The proposed method is implemented in a small quadcopter, and the results show that our method can achieve the lowest position error (0.11m) and a comparable velocity error (0.23m/s) across the benchmarking algorithms running on the robot's onboard computer. The flight experiments prove that the tracking results from the proposed method can make the robot efficiently alter its trajectory for navigating dynamic environments. Our software is available on GitHub\footnote{\url{https://github.com/Zhefan-Xu/onboard_detector}} as an open-source ROS package.

.
\end{abstract}

\section{Introduction}
\IEEEPARstart{S}{mall} autonomous mobile robots, frequently employed in indoor scenarios, often operate in dynamic and unpredictable environments populated by humans, vehicles, and other robots. Ensuring safe navigation in such settings necessitates real-time, accurate perception of dynamic obstacles. However, many small robots are only equipped with onboard computers with limited computational capabilities and rely on RGB-D cameras. This makes GPU-intensive learning-based methods, common in autonomous driving, unsuitable. Hence, the development of a lightweight RGB-D camera-based dynamic obstacle detection and tracking method is necessary for autonomous robots operating in dynamic environments.
\begin{figure}[t] 
    \vspace{0.10cm}
    \centering
    \includegraphics[scale=1.13]{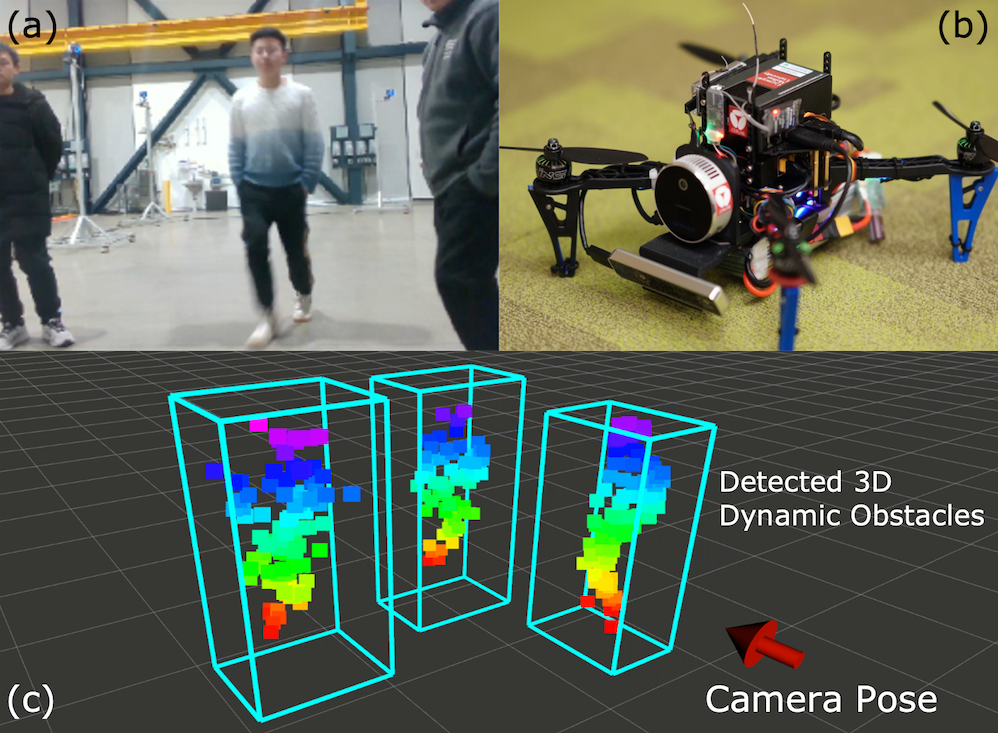}
    \caption{The onboard dynamic obstacle detection results from the proposed DODT algorithm. (a) The camera RGB view. (b) An example of an autonomous robot with an RGB-D camera. (c) The onboard 3D dynamic obstacle detection results shown as blue bounding boxes with point clouds.}
    \label{intro figure}
\end{figure}

There are three challenges in small mobile robots' detection and tracking. First, small mobile robots' onboard computation resources are limited, making GPU-demanding learning-based methods \cite{second}\cite{pvrcnn} not applicable. Note that we define small mobile robots as those with weights below 1.5kg, equipped with low-power (10-20Watts) onboard computers measuring around 10cm in length, shown in Fig. \ref{intro figure}b. Second, the range and field of view (FOV) of depth cameras suited for small mobile robots are limited, which makes obstacles either too close or too far and thus not detectable. For example, the ideal depth range of the popular Intel RealSense D435i depth camera is from 0.3m to 3.0m. This camera limitation makes some previous works \cite{ETH_pointcloud}\cite{zju_pointcloud} only capable of tracking obstacles in the short range. Third, the noises from the depth value estimation of the camera are not negligible, especially for those noise-sensitive non-learning methods \cite{hku_pointcloud}\cite{xu2022real}. The camera noises can make the detection algorithm not only estimate obstacle states inaccurately but also produce high-frequency false-positive and false-negative results, leading to confusion for obstacle avoidance planners.

To solve these issues, this paper presents an onboard 3D dynamic obstacle detection and tracking (DODT) method based on an RGB-D camera. In contrast to other low-computational algorithms \cite{ETH_pointcloud}\cite{zju_pointcloud}\cite{xu2022real}, which employ a single detector, we propose a novel ensemble detection strategy combining multiple computationally efficient but low-accuracy detectors to obtain fast and more accurate obstacle detection results. Moreover, the proposed method incorporates feature-based data association, utilizing statistical features from point clouds, and employs the Kalman filter for obstacle tracking. This approach reduces tracking mismatches that can occur with the center-distance-based association methods employed by benchmarking algorithms. Then, we use both point cloud and velocity criteria to identify dynamic obstacles. Finally, the system introduces a novel usage for the learning-based detector as an auxiliary and optional module to enhance the detection range and dynamic obstacle identification when the robot's computation resources are enough. The  contributions of this work are:
\begin{itemize}
    \item \textbf{Efficient Ensemble Detection:} Different from other detection and tracking algorithms designed for low-computational robots with a single detector, the proposed algorithm runs multiple computationally efficient and low-accuracy detectors with a novel ensemble strategy to obtain more accurate results with high efficiency. 
    \item \textbf{Feature-based Association and Tracking:} Unlike the traditional center-distance-based association methods in other algorithms, the feature-based association reduces tracking mismatches by utilizing statistical features from point clouds, improving the tracking accuracy.
    \item \textbf{Auxiliary Learning-based Detection Module:} The system incorporates a novel integration of the learning-based detector as an auxiliary module, enhancing the detection range and dynamic obstacle identification when the robot has sufficient computational resources.
\end{itemize}

\section{Related Work}
Among small robots with limited computational power, such as UAVs and small UGVs, the detection and tracking methods can be categorized based on the sensors, including LiDARs \cite{lidar1}\cite{lidar2}\cite{lidar3}\cite{lidar4}, event cameras \cite{event1}\cite{event2}, and RGB-D cameras \cite{ETH_pointcloud}\cite{zju_pointcloud}\cite{xu2022real}. Among them, the RGB-D camera is one of the most popular sensors for small mobile robots, and there are mainly two ways of using the RGB-D camera.

\textbf{Image-based methods:} Most methods in this category leverage depth images for 3D obstacle detection. For instance, in \cite{ETH_uv_15}, depth images are employed to generate U-depth maps and V-depth maps, enabling the estimation of obstacle states and proving safe navigation with static obstacles. Building on this, Lin et al. \cite{delft_u_map} adopt a similar U-depth map to detect and track obstacles, representing them as 3D ellipsoids. To enhance the accuracy of obstacle dimension estimation, the restricted V-depth map is introduced in \cite{restricted_v_map}. In \cite{xu2022real}, dynamic obstacles identified from the depth and U-depth maps are characterized by their estimated velocities. These dynamic obstacle detection outcomes are integrated with the occupancy map to navigate dynamic environments. In contrast to prior depth image-based approaches, Lu et al. \cite{hku_yolo} apply the YOLO detector to effectively avoid fast and small dynamic obstacles. Additionally, Sun et al. \cite{motion_removal} employ image differences to identify all dynamic points from RGB images. Moreover, Logoglu et al. \cite{3_img_diff} combine the 3-image-difference technique with epipolar constraints to determine dynamic obstacles. They extend their approach by utilizing scene flow, an extension of optical flow, in \cite{poitnPWC} \cite{optical_flow}, for detecting the velocity of each pixel and identifying dynamic points. Some alternative methods focus on detecting and segmenting dynamic obstacles in 2D image planes to enhance SLAM robustness. In \cite{motion_removal}\cite{ds_slam}\cite{Dai_remove}\cite{DOT}, these approaches concentrate on removing dynamic obstacles from images to mitigate estimation errors, while Qiu et al. \cite{airdos} detect pedestrian skeletons to improve SLAM optimization.

\textbf{Point cloud-based methods:} Unlike image-based methods, point cloud-based approaches directly detect 3D obstacles using geometric information from point clouds. In \cite{ETH_pointcloud}, a point cloud clustering method is combined with the YOLO detector for human detection. Wang et al. \cite{zju_pointcloud} employ a similar clustering-based detection approach for indoor dynamic obstacle avoidance using a quadcopter. To enhance obstacle tracking robustness, Chen et al. \cite{hku_pointcloud} propose using point cloud feature vectors and object track points to identify correct object matches and estimate their states. In \cite{hku_lidar_avoidance}, a KD-Tree map is directly constructed from the LiDAR point cloud for dynamic obstacle avoidance. Min et al. \cite{kernel_method} represent dynamic obstacles in a dynamic occupancy map and employ kernel inference to reduce computation. Likewise, in \cite{dsp_map}, a dual-structure particle-based dynamic occupancy map is utilized to represent dynamic environments and classify obstacle particles as static or dynamic.


Both image and point cloud methods can suffer from misdetection due to noise and complex environments. To address this, we propose an ensemble method that leverages different detectors to mitigate their individual shortcomings. Additionally, we suggest using the learning-based method as an optional auxiliary module, enhancing adaptability for robots with varying computational resources.

\section{Methodology} \label{method}
\begin{figure*}[t] 
    \centering
    \includegraphics[scale=0.46]{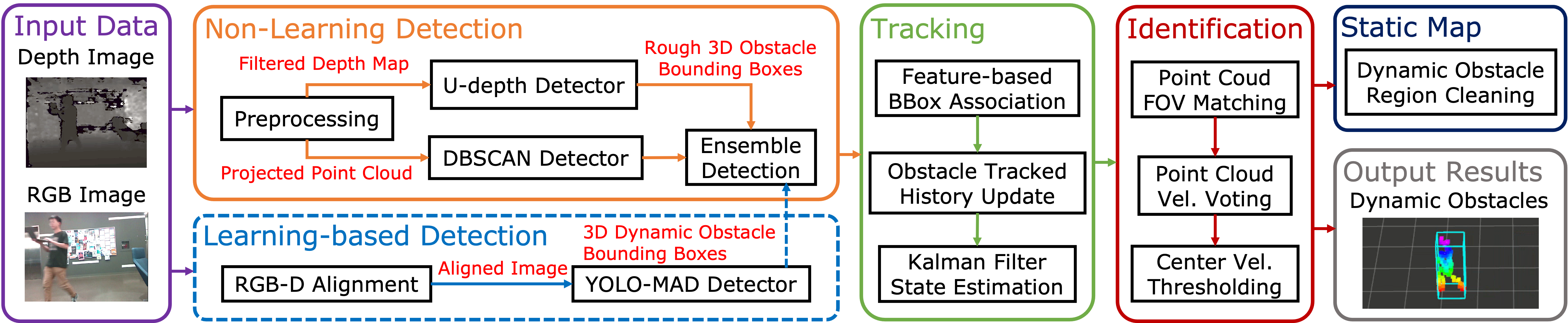}
    \caption{The proposed dynamic obstacle detection and tracking system (DODT) framework. The input data are the RGB-D images. The non-learning detection module first uses the depth image to detect generic obstacles. Then, the tracking module is applied to track and estimate the obstacles states. With the identification module, the dynamic obstacles are identified from all detected obstacles. Finally, the output results show the dynamic obstacles' bounding boxes. The dynamic obstacle regions are cleaned in the static occupancy map. The optional learning-based detection module, presented in the blue dotted line, uses color and depth images to detect dynamic obstacles, enhancing the detection range and dynamic obstacle identification. }
    \label{system_framework}
\end{figure*}
\subsection{System Overview}
Considering the payload and computational constraints of small mobile robots, both computational-intensive learning-based 3D object detectors and heavy LiDAR systems become impractical. To address this constraint, we have devised a lightweight detection and tracking framework comprising three core modules: the detection module, the tracking module, and the identification module, as shown in Fig. \ref{system_framework}. The detection module comprises a non-learning and a learning-based component. The non-learning part employs depth images and two non-learning detectors for generic obstacle detection. Meanwhile, the learning-based module uses aligned RGB-D images for direct dynamic obstacle detection, and its results are combined with the non-learning module. Details of each detector are in Sec. \ref{detectors section}, with ensemble detection explained in Sec. \ref{ensemble detection section}. Refined 3D bounding boxes are used in the tracking module (Sec. \ref{tracking section}) to estimate obstacle states using historical data. The identification module (Sec. \ref{identification section}) classifies obstacles as static or dynamic based on state and tracking history. The system outputs dynamic obstacle bounding boxes, and dynamic obstacle regions are cleared in the static map for navigation.

\subsection{3D-Obstacle Detectors} \label{detectors section}
This section introduces three computationally efficient but low-accuracy 3D obstacle detectors: the U-depth, the DBSCAN, and the YOLO-MAD detector. Note that all detection results are represented as axis-aligned bounding boxes. We select the U-depth and the DBSCAN detectors as the non-learning detectors due to their high computational efficiency demonstrated in small UAV 3D dynamic obstacle detection applications \cite{zju_pointcloud}\cite{xu2022real}. Besides, their detection errors come from different sources (the depth image and the point cloud) obtained from the RGB-D camera, ensuring the ensemble strategy takes effect. For the learning-based detector, we choose an extremely lightweight implementation of a popular model and extend it into a 3D detector, which can run in real time on onboard computers without GPU acceleration.

\begin{figure}[t] 
    \centering
    \includegraphics[scale=0.57]{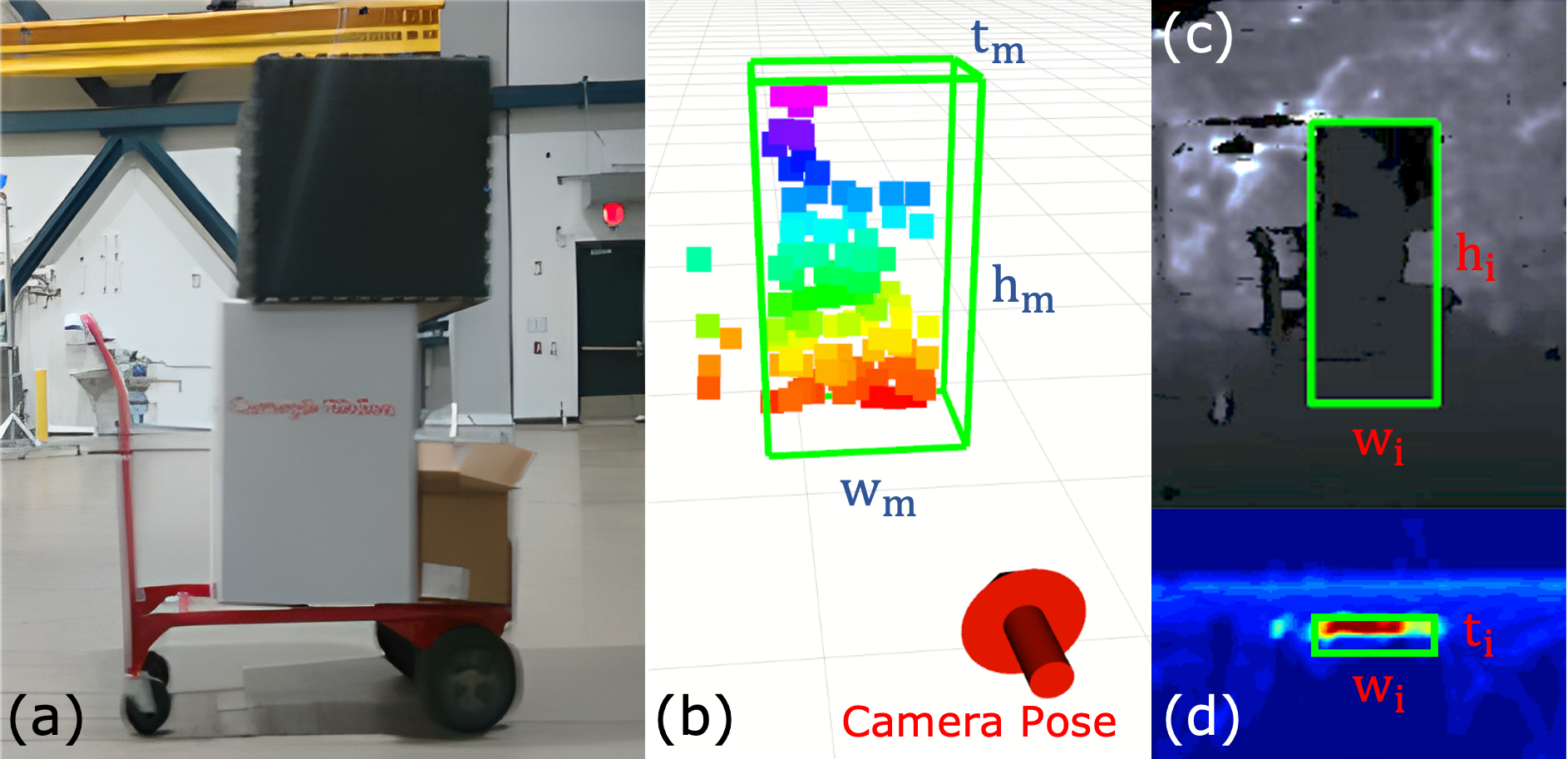}
    \caption{Illustration of the U-depth detector. (a) The camera RGB view. (b) The detected 3D bounding box with the obstacle point cloud. (c) The 2D detection on the depth map. (d) The 2D detection on the U-depth map.}
    \label{u_depth_detector}
\end{figure}

\textbf{U-depth Detector:} The U-depth detector for obstacle detection is mentioned in the previous works \cite{ETH_uv_15}\cite{delft_u_map}\cite{xu2022real}. Overall, the detector takes the depth image to generate 3D bounding boxes of static and dynamic obstacles. Fig. \ref{u_depth_detector} visualizes sample detection results. There are three steps in the U-depth detector: (1) the U-depth map generation, (2) the line grouping on U-depth, and (3) the depth continuity search on the original depth image. 

The U-depth map can be intuitively viewed as the top-down view from the camera. It has the same width as the original depth image, and its vertical axis from top to bottom indicates the increasing distance to the camera. When we get a depth image, we can compute the U-depth map using the column depth value histogram. Fig. \ref{u_depth_detector}c and Fig. \ref{u_depth_detector}d show a depth image and U-depth map pair. Then, we can perform the line grouping method on the generated U-depth map to get the 2D bounding box of the obstacle of width $\text{w}_{\text{i}}$ and thickness $\text{t}_{\text{i}}$ shown in Fig. \ref{u_depth_detector}d (note that $\text{i}$ indicates the image plane). With the obstacle width $\text{w}_{\text{i}}$, we do the depth value continuity check on the original depth image to get the height $\text{h}_{\text{i}}$ of the obstacle shown in Fig. \ref{u_depth_detector}c. After having both 2D bounding boxes in the U-depth map and the original depth image, we can triangulate 3D points into the camera frame and perform coordinate transform to get the obstacle position and dimension of the world/map coordinate frame (Fig. \ref{u_depth_detector}b).

\begin{figure}[t] 
    \centering
    \includegraphics[scale=0.43]{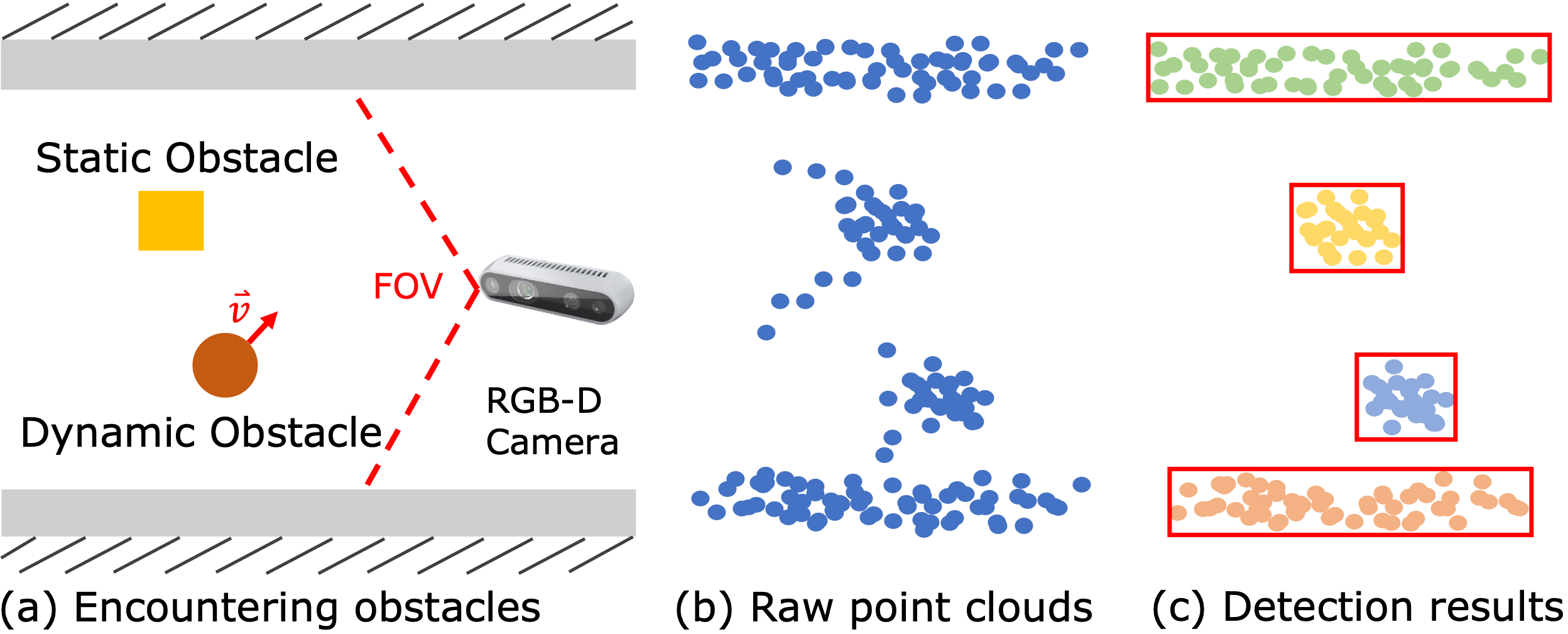}
    \caption{Illustration of the DBSCAN detector. (a) The robot encounters obstacles in a corridor. (b) The raw point cloud data from the RGB-D camera are unstructured and noisy. (c) The DBSCAN detector takes the filtered point cloud and performs clustering to get obstacles' bounding boxes.}
    \label{db_scan_detector}
\end{figure}

\textbf{DBSCAN Detector:} Unlike the image-based detector, the DBSCAN detector uses point cloud data to detect obstacles. DBSCAN is an unsupervised machine-learning algorithm for clustering which can automatically determine the cluster number. The illustration of the DBSCAN detector is shown in Fig. \ref{db_scan_detector}. When the robot encounters obstacles, the raw point cloud data can be triangulated from the depth image as shown in Fig. \ref{db_scan_detector}b. Note that because of the sensor, the point cloud data can be noisy on the obstacle boundaries. So, we apply the voxel filter proposed in \cite{ETH_pointcloud} to remove the noise of the point cloud and then perform DBSCAN clustering to get obstacles' bounding boxes (Fig. \ref{db_scan_detector}c). Similar to the U-depth detector, the DBSCAN detector does not need a training dataset and only requires a few computation resources.

\begin{figure}[t] 
    \centering
    \includegraphics[scale=0.66]{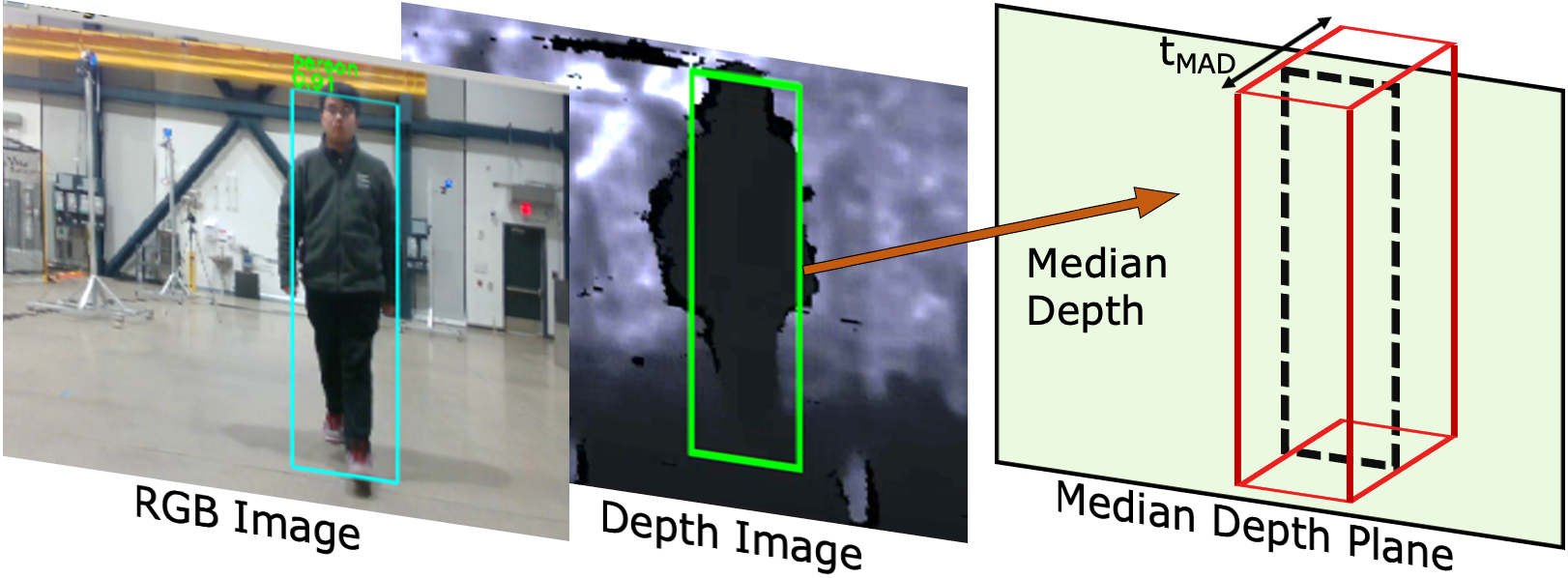}
    \caption{Illustration of the YOLO-MAD detector. The RGB image is used to get the 2D detection result, and then the bounding box on the depth image is obtained. With the 2D result on the depth image, the 3D bounding box is calculated by the proposed median absolute deviation (MAD) method. }
    \label{yolo_mad_detector}
\end{figure}

\textbf{YOLO-MAD Detector:}
The previously mentioned detectors rely on geometric structures of either depth images or point clouds. So, they cannot identify the type of obstacles (i.e., static or dynamic) and might even fail when the obstacles are far from the camera. To overcome these limitations, we introduce our 3D YOLO-MAD detector based on the 2D YOLOFastestDet, which can run real-time at an onboard CPU such as Intel NUC. The illustration of the YOLO-MAD detector is shown in Fig. \ref{yolo_mad_detector}. The detector first detects the 2D bounding box of each obstacle on the RGB image and finds the corresponding region on the aligned depth image. To find the depth and thickness of the 2D bounding box, we first calculate the median absolute deviation (MAD) based on the median depth value $\Tilde{d}$ in the bounding box region $\mathcal{R}_{\text{box}}$:
\begin{equation}
  \text{MAD} = \textbf{median}(|d_{i} - \Tilde{d}|), \ \ d_{i} \in \textbf{depth}(\mathcal{R}_{\text{box}}),
\end{equation}
where $d_{i}$ is the depth value of $i$th pixel in the bounding box region $\mathcal{R}_{\text{box}}$. Then, we can search the minimum depth $d_{\text{min}}$ and maximum depth $d_{\text{max}}$ in the MAD range $\mathcal{S}_{\text{MAD}}$:
\begin{equation}
    \mathcal{S}_{\text{MAD}} = \{d_{i} | \Tilde{d} - n \cdot \text{MAD} \leq d_{i} \leq \Tilde{d} + n \cdot \text{MAD} \},
\end{equation}
where $n$ is a user-defined parameter. The obstacle's thickness $t_{\text{MAD}}$ can be calculated based on the minimum and maximum depth values. The MAD range $\mathcal{S}_{\text{MAD}}$ can help filter the outlier depth values in the bounding box region from the background and the sensor noises. Finally, we can triangulate the points from the depth image at the median depth plane with the thickness to get the 3D obstacle's bounding box. Since this learning-based detector can still be computationally heavy for some extremely low-power onboard computers, we treat it as an optional and auxiliary module in our framework.

\subsection{Ensemble Detection} \label{ensemble detection section}
This section introduces our proposed ensemble detection method to obtain refined obstacles' bounding boxes. In our framework, three detectors run in parallel and individually detect obstacles' bounding boxes. Since the previously mentioned detectors are designed to compensate for the detection accuracy for high-speed performance, they are all sensitive to different environments and sensor noises, leading to false positives and inaccurate obstacle dimension estimation. So, the intuition of the ensemble detection is to combine the detection results of different detectors and find their \say{mutual agreements} of detection results for reducing the noise effects. This technique can significantly improve detection robustness and accuracy with environment and sensor noises.

The proposed ensemble detection algorithm follows a pairwise manner presented in Alg. \ref{ensemble detection algorithm}. When we obtain two sources of detection results, we go through each bounding box $\text{b}_{\text{d1}}$ from one detector's results (Line \ref{for loop}). For the bounding box $\text{b}_{\text{d1}}$, the algorithm finds the bounding box $\text{b}_\text{match1}$ with the highest intersection-over-union (IOU) score from the other detection bounding boxes (Line \ref{IOU1}). Following the same way, the bounding box $\text{b}_\text{match2}$ is obtained by finding the highest IOU match of $\text{b}_\text{match1}$ in the first detection bounding boxes (Line \ref{IOU2}). Through this process, we want to find the bounding boxes that are detected by both detectors. Then, we need to ensure that the IOU score of their matched bounding boxes exceeds the predefined threshold and that their matched bounding boxes have the highest IOU score to each other (Line \ref{match condition}). Finally, we fuse two bounding boxes into a new ensembled bounding box (Lines \ref{fuse bbox}-\ref{add ensemble bbox}). We adopt a conservative method for fusing bounding boxes: the new ensembled bounding box takes the maximum values in dimensions and the average value in positions. In our system framework (Fig. \ref{system_framework}), we first ensemble detection results from the U-depth and DBSCAN detectors and then combine the YOLO-MAD results if the learning-based module is running.

\begin{algorithm}[t] \label{ensemble detection algorithm}
\caption{Ensemble Detection Algorithm} 
\SetAlgoNoLine%
$\mathcal{B}_{en} \gets  \emptyset$ \Comment*[r]{ensembled bounding boxes}
$\mathcal{B}_{d1} \gets  \textbf{getDetBBox1}()$ \Comment*[r]{detector1 results}
$\mathcal{B}_{d2} \gets  \textbf{getDetBBox2}()$ \Comment*[r]{detector2 results}
\For{$\normalfont{\text{b}}_{\normalfont{\text{d1}}}$ \normalfont{\textbf{in}} $\mathcal{B}_{\normalfont{\text{d1}}}$}{  \label{for loop}
     $\mathcal{S}_{\text{iou1}}, \normalfont{\text{b}}_{\normalfont{\text{match1}}} \gets \textbf{findBestIOUMatch}(\normalfont{\text{b}}_{\normalfont{\text{d1}}}, \mathcal{B}_{d2})$\; \label{IOU1} 
     $\mathcal{S}_{\text{iou2}}, \normalfont{\text{b}}_{\normalfont{\text{match2}}} \gets \textbf{findBestIOUMatch}(\normalfont{\text{b}}_{\normalfont{\text{match1}}}, \mathcal{B}_{d1})$\; \label{IOU2}
     $\mathcal{C}_{\text{match}} \gets \normalfont{\text{b}}_{\normalfont{\text{match2}}}$ \normalfont{\textbf{is}} $\normalfont{\text{b}}_{\normalfont{\text{d1}}}$\;
     \If{$\mathcal{S}_{\normalfont{\text{iou1}}} > \mathcal{S}_{\normalfont{\text{thr}}}$ \normalfont{\textbf{and}} $\mathcal{S}_{\normalfont{\text{iou2}}} > \mathcal{S}_{\normalfont{\text{thr}}}$ \normalfont{\textbf{and}} $\mathcal{C}_{\normalfont{\text{match}}}$}{ \label{match condition}
        $\text{b}_{\text{en}} \gets \textbf{fuseBBoxes}(\normalfont{\text{b}}_{\normalfont{\text{d1}}}, \normalfont{\text{b}}_{\normalfont{\text{match1}}})$\; \label{fuse bbox}
        $\mathcal{B}_{en}.\textbf{push\_back}(\text{b}_{\text{en}})$\; \label{add ensemble bbox}
    }
}
$\textbf{return} \ \mathcal{B}_{en}$\;
\end{algorithm}

\subsection{Data Association and Tracking} \label{tracking section}
Overall, the proposed module first applies the feature-based data association method to match the detected obstacles at the current time $t_{n}$ with the obstacles at the previous time $t_{n-1}$. Then, it applies the Kalman filter with the constant-acceleration motion model to estimate the obstacles' states and add them to the estimation histories. In contrast to the constant-velocity model used in previous works \cite{delft_u_map}\cite{zju_pointcloud}\cite{xu2022real}, the constant acceleration model offers more accurate state estimation and dynamic obstacle identification.

\textbf{Feature-based Data Association}:
The detected obstacles at the current time $t_{n}$ are associated with the obstacles at the previous time $t_{n-1}$ using the feature comparison. The feature vector of the obstacle $O_{i}$ is defined as:
\begin{equation}
    feat(O_{i}) = [pos(i), dim(i), len(i), std(i)],
\end{equation}
where $pos(i)$ is the obstacle's center position, $dim(i)$ is the obstacle's dimension in x, y and z direction, $len(i)$ is the obstacle's point cloud size, and $std(i)$ is the obstacle's point cloud standard deviation. Then, we perform normalization for the feature vector to reduce the effects from the different dimensions. After that, the similarity score between obstacles $O_{i}$ and $O_{j}$ is calculated using the following equation:
\begin{equation}
    sim(O_{i}, O_{j}) = exp(-||feat(O_{i})-feat(O_{j}))||_{2}^{2}),
\end{equation}
where we take the exponential of the negative L2 norm of the feature difference. With the scores, the obstacle $O_{i}^{t_{n}}$ at the current time $t_{n}$ can be matched with the obstacle $O_{j}^{t_{n-1}}$ at the previous time $t_{n-1}$ with the highest similarity score $sim_{\text{max}}$. Instead of directly using the previous obstacle's feature, we apply the linear propagation to get the predicted obstacle's position and replace the previous obstacle's position with the predicted position in the feature vector. Also, the highest similarity score must be higher than a predefined threshold ($sim_{\text{max}} > T_{sim}$) to prevent incorrect associations.

\begin{figure}[t] 
    \centering
    \includegraphics[scale=1.1]{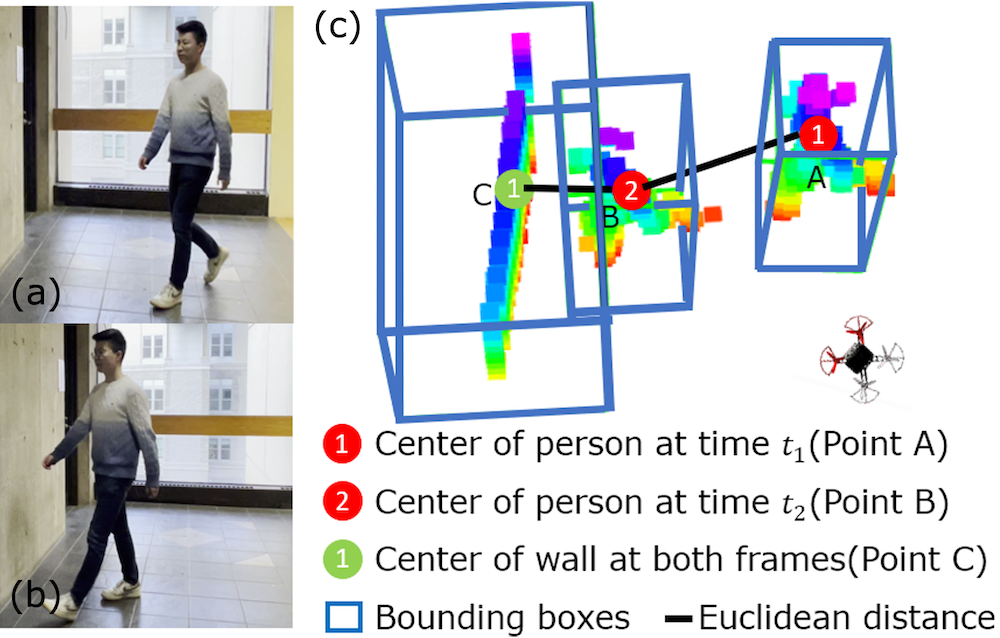}
    \caption{Illustration of the issue with the center-distance-based data association method. (a) The RGB image at time $t_1$. (b) The RGB image at time $t_2$. (c) The center-distance-based data association method might fail by incorrectly associating the current detected person with the wall. }
    \label{data_association}
\end{figure}

The proposed feature-based data association method can overcome the drawback of traditional center-distance-based association, as shown in Fig. \ref{data_association}. In Fig. \ref{data_association}a and b, a scenario is presented where a person approaches the wall with the point clouds of all obstacles shown in Fig. \ref{data_association}c. Since the center of the wall (Point C) is closer to the person's position at the current time $t_{2}$ (Point B) than the person's position at the previous time $t_{1}$ (Point A), a center-distance-based tracking will associate the person with the wall. On the contrary, if the proposed feature-based association method is applied, the person and wall will not be matched together because of the obvious differences in the obstacles' dimensions, velocities, point cloud sizes, and standard deviations. So, the detected person at the current time $t_{2}$ will be correctly associated with the person at the previous time $t_{1}$.


\textbf{Constant-Acceleration Kalman Filter}: The states of each obstacle are estimated by the Kalman filter with a constant-acceleration motion model. Unlike the previous work \cite{zju_pointcloud} \cite{xu2022real}, where the velocities of obstacles are assumed to be constant, our method allows the obstacles' velocities to change without increasing the complexity of the motion model too much. We will discuss all quantities in global map frame for simplicity. The obstacle states are defined as $X=[x, y, \Dot{x} , \Dot{y}, \Ddot{x} , \Ddot{y}]^{T}$, including the position, the velocity, and the acceleration in x and y directions. The measurement vector is the same as the obstacle state vector. To calculate the measurement of the velocity vector $\textbf{V}_{i}$ and acceleration vector $\textbf{A}_{i}$ at time $t$, we adopt the following equations:
    \begin{equation}
    \textbf{V}_{t} = \frac{\textbf{P}_{t} - \textbf{P}_{t-1}}{\delta t}, \ \textbf{A}_{t} = \frac{\textbf{V}_{t} - \textbf{V}_{t-1}}{\delta t}, 
\end{equation}
where $\delta t$ is the time difference. Note that we take the data from several time differences $\delta t$ to calculate smoother observations. In this way, the system model is described by:
\begin{equation}
    X_{t|t-1} = AX_{t-1}+Bu_{t-1} +Q,
\end{equation}
where $A$ is the state transition matrix, $Q$ is the covariance of the motion model noise, $u$ is the control input, which is zero in this case. Since the acceleration model is assumed, the state transition matrix can be calculated by:
\begin{equation}
A = 
\begin{bmatrix}
1 & 0 & \delta t & 0 & \frac{\delta t^2}{2} & 0\\
0 & 1 & 0 & \delta t & 0 & \frac{\delta t^2}{2}\\
0 & 0 & 1 & 0 & \delta t & 0\\
0 & 0 & 0 & 1 & 0 & \delta t\\
0 & 0 & 0 & 0 & 1 & 0\\
0 & 0 & 0 & 0 & 0 & 1\\
\end{bmatrix},
\end{equation}
and the system measurement is defined as:

\begin{equation}
    Z_{t} = HX_{t} + R,
\end{equation}
where the measurement matrix $H$ is an identity matrix, and $R$ is the covariance of measurement noise.

\subsection{Dynamic Obstacle Identification} \label{identification section}
\begin{figure}[t] 
    \centering
    \includegraphics[scale=0.33]{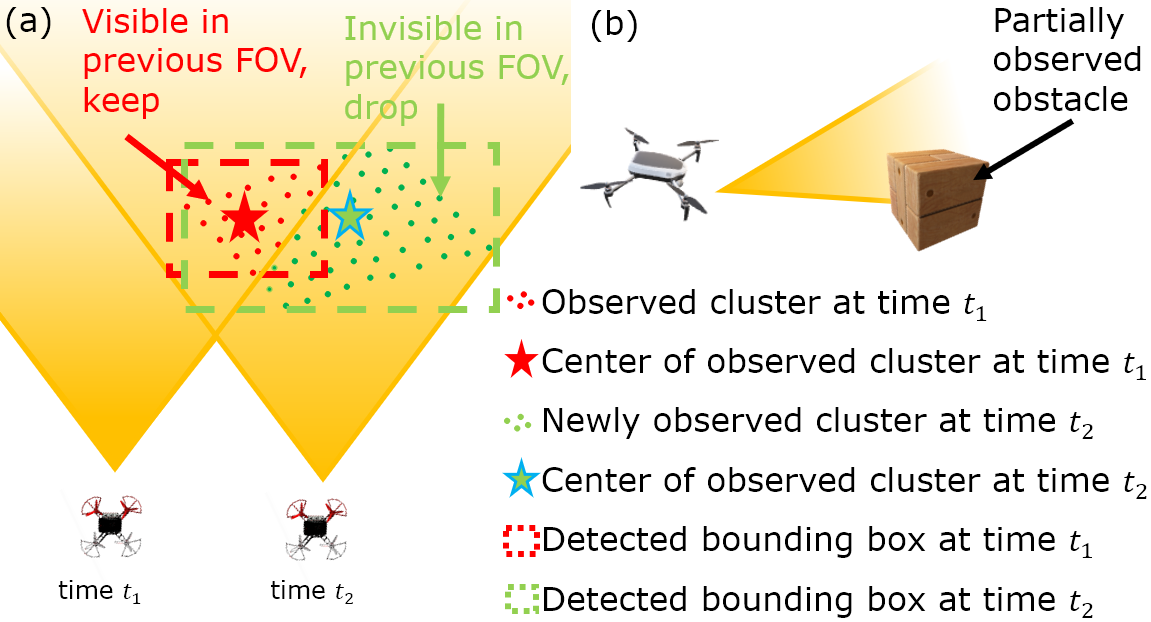}
    \caption{Illustration of removing the invalid points using the field of view (FOV) criteria. (a) The analysis of the observed obstacle's point cloud at different time. (b) The robot detecting a partially visible obstacle. }
    \label{is_in_fov}
\end{figure}
This section describes how to identify the status of an obstacle (dynamic or static). By default, any quantities defined in the following are at the current time $t_{n}$. As the first dynamic obstacle identification criteria, all the bounding boxes of obstacles with the center velocity $\textbf V_{center}$ less than a threshold $T_{vel}$ will be classified as static. Although the velocity criteria should theoretically filter out all static obstacles, the noises from detection and state estimation can cause false-positive dynamic obstacle identification. To reduce the false-positive identification results, in the second identification step, the module takes all valid points of an obstacle's point cloud to vote for its status. In this step, every point at the current time $t_{n}$ is matched with its corresponding point at the time $t_{n-k}$ by the nearest neighbor search. After determining the correspondence, the velocity of each point $\textbf V_{vote}^i$ is calculated. Then, a point will vote for the obstacle as dynamic if its velocity exceeds a predefined threshold $T_{vote}$. If the ratio of dynamic votes $N_{vote}$ over the number of valid points $N_{valid}$ is higher than another threshold $T_{ratio}$, the obstacle will be identified as a dynamic obstacle:
\begin{equation}
    \frac{N_{vote}}{N_{valid}} > T_{ratio}.
\end{equation}

Before the dynamic voting process, it is necessary to drop the invalid points from the point cloud. First, if any point $p_{i,j}$ with the point cloud index $i$ in obstacle $j$ has an invalid velocity $\textbf V_{vote}^{i}$, it will be removed from the dynamic voting process. The valid velocity should satisfy the condition:
\begin{equation}
    angle(\textbf V_{vote}^{i},\textbf V_{center}^j) < \frac{\pi}{2},
\end{equation}
where we ensure that points with incorrect velocity estimations are removed. Second, if any point $p_{i,j}$ at time $t_{n}$ is invisible at time $t_{n-k}$, it will also be removed from voting shown in Fig. \ref{is_in_fov}. Fig. \ref{is_in_fov}(b) shows a scenario where a robot approaches a partially visible static obstacle. At the previous time $t_{1}$, only red points are visible; the detected center of the obstacle is the red star. At the current time $t_{2}$, the whole box is visible, and the center of the obstacle shifts a lot. In this case, the obstacle will have a large center velocity $\textbf V_{center}$ and voting velocity $\textbf V_{vote}$ due to incorrect points correspondence. Our method drops the newly observed points from the voting and identifies the obstacle as static. Finally, when the YOLO-MAD Detector is applied, its classification results will be used for dynamic obstacle identification, skipping all the processes mentioned above.

\section{Result and Discussion}
To evaluate the performance of the proposed method, we conduct experiments in dynamic environments. The algorithm is implemented in C++, running on two customized quadcopters with the Intel NUC and NVIDIA Jetson Xavier NX onboard computers, respectively. All the computations are performed real-time on the robots' onboard computers.

\subsection{Performance Benchmarking}

To assess our algorithm's performance, we conduct comparative experiments with state-of-the-art dynamic obstacle detection and tracking methods in the UAV platform \cite{delft_u_map}\cite{zju_pointcloud}\cite{xu2022real}. We also evaluate the impact of ensemble detection and feature-based association and tracking by comparing our method's performance with and without these features. For experiments without ensemble detection (DDOT w/o Ens), we use the U-depth detector due to its higher accuracy compared to the DBSCAN detector. In the absence of feature-based association and tracking (DODT w/o FAT), we apply center-distance-based association with the constant-velocity model. We employ various evaluation metrics, including position and velocity estimation errors and the false-positive detection rate. The false-positive rate is determined by dividing the number of misdetections (identifying static obstacles as dynamic) by the total number of detections. Ground truth measurements are acquired from the OptiTrack motion capture system. Table \ref{detection_result} summarizes the comparison results. Our DODT method exhibits the lowest position errors among all methods, with our velocity error ranking second, comparable to Method III \cite{xu2022real}. Ensemble detection significantly reduces the false-positive detection rate by leveraging consensus among detectors and enhances obstacle position and velocity estimation accuracy. Additionally, the feature-based association and tracking method results in lower state estimation errors and a reduction in false-positive rates. From the experiment observation, this reduction in the state estimation errors and false-positive rates comes from fewer obstacle mismatches and more accurate velocity estimation. 

\begin{table}[t]
\begin{center}
\caption{Benchmarking of the detection and tracking results in the position errors, velocity errors and the false positive rates.} \label{detection_result}
\begin{tabular}{ |c | c | c | c| } 
 \hline

  Method & Pos. Err. (m)  & Vel. Err. (m/s) & FP Rate (\%) \Tstrut\\ 
 \hline

 Method I \cite{delft_u_map}  & $0.28$  & $0.47$ & N/A \Tstrut\\ 
 \hline

 Method II \cite{zju_pointcloud} & $0.18$  & $0.29$ & $16.4\%$ \Tstrut\\  
 \hline
 
 Method III \cite{xu2022real} & $0.19$  & $\textbf{0.21}$ & $19.6\%$ \Tstrut\\  
 \hline

 DODT w/o Ens & $0.17$  & $0.30$ & $18.6\%$ \Tstrut\\  
 \hline   
 DODT w/o FAT & $0.14$  & $0.29$ & $6.5\%$ \Tstrut\\
 \hline
 \textbf{DODT (Ours)} & $\textbf{0.11}$  & $0.23$ & $\textbf{3.7\%}$ \Tstrut\\  
 \hline
\end{tabular}
\end{center}
\end{table}

\begin{figure}[t] 
    \centering
    \includegraphics[scale=1.11]{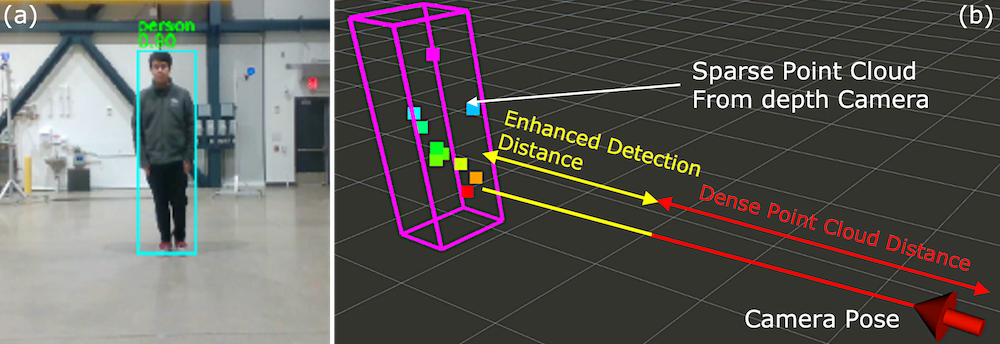}
    \caption{Illustration of enhancing detection range by the auxiliary learning-based module. The red line measures the maximum ideal range to produce dense point cloud data for the DBSCAN and U-depth detectors to detect obstacles. The yellow line indicates the increased detection distance.}
    \label{YOLO demo}
\end{figure}

The result illustration of enhancing detection range by the auxiliary learning-based module is visualized in Fig. \ref{YOLO demo}. In Fig. \ref{YOLO demo}b, we label our depth camera's dense point cloud distance (around 3m). Since both non-learning detectors, the U-depth and the DBSCAN detectors, require geometric information from either depth image or point cloud, detecting obstacles using the non-learning detectors outside the dense point cloud region can fail. On the contrary, the learning-based module can use the color image to detect obstacles (Fig. \ref{YOLO demo}a) even though the obstacle is in a sparse point cloud region. Fig. \ref{YOLO demo}b shows that our YOLO-MAD detector can successfully detect the dynamic obstacle (shown as the purple bounding box) in the sparse point cloud region with the increasing detection distance labeled as the yellow line. 

\subsection{Runtime Analysis}
The runtime of the entire system is detailed in Table \ref{runtime table}, with measurements conducted on both the Intel NUC and Xavier NX onboard computers. Notably, the total runtime for the Intel NUC and Xavier NX is 19.12ms and 40.08ms, respectively, indicating the real-time performance on both platforms. The runtime breakdown results reveal that the YOLO-MAD detector consumes a significant portion of the processing time, accounting for 75.7\% and 59.5\% of the total detector runtime on the Intel NUC and Xavier NX, respectively. As discussed in Section \ref{detectors section}, we recommend using the YOLO-MAD detector as an optional and auxiliary module based on the computational resources. Experiments demonstrate that if the user disables the learning-based module, the detection frame rate on the Intel NUC and Xavier NX can increase substantially, reaching around 210Hz and 60Hz, respectively, up from 50Hz and 25Hz.
\begin{table}[t]
    \centering
    \caption{The runtime of each module of the proposed system.}
    \begin{tabular}{ l c c } 
    \hline
    System Modules & Intel NUC (ms) & Xavier NX (ms) \Tstrut\\
    \hline
    U-depth detection & 3.4 & 12.0\Tstrut\\ 
    DBSCAN detection & 1.3 & 4.0\Tstrut\\ 
    YOLO-MAD detection & 14.3 & 23.5\Tstrut\\ 
    Feature-based Data Assoc. & 0.03 & 0.08\Tstrut\\ 
    Kalman filter tracking & 0.07 & 0.17\Tstrut\\ 
    Dynamic Obstacle Id. & 0.12 & 0.33\Tstrut\\ 
    \textbf{System Total Runtime} & \textbf{19.12} & \textbf{40.08} \Tstrut\\ 
    \hline

    \end{tabular}
    \label{runtime table}
\end{table}


\subsection{Physical Experiments}
To verify the proposed algorithm's performance in robot navigation, we conduct handheld experiments using the robot camera and do the autonomous navigation tests with the trajectory planner \cite{xu2022vision}\cite{xu2022dpmpc} in dynamic environments. 

\begin{figure*}[t] 
    \centering
    \includegraphics[scale=1.17]{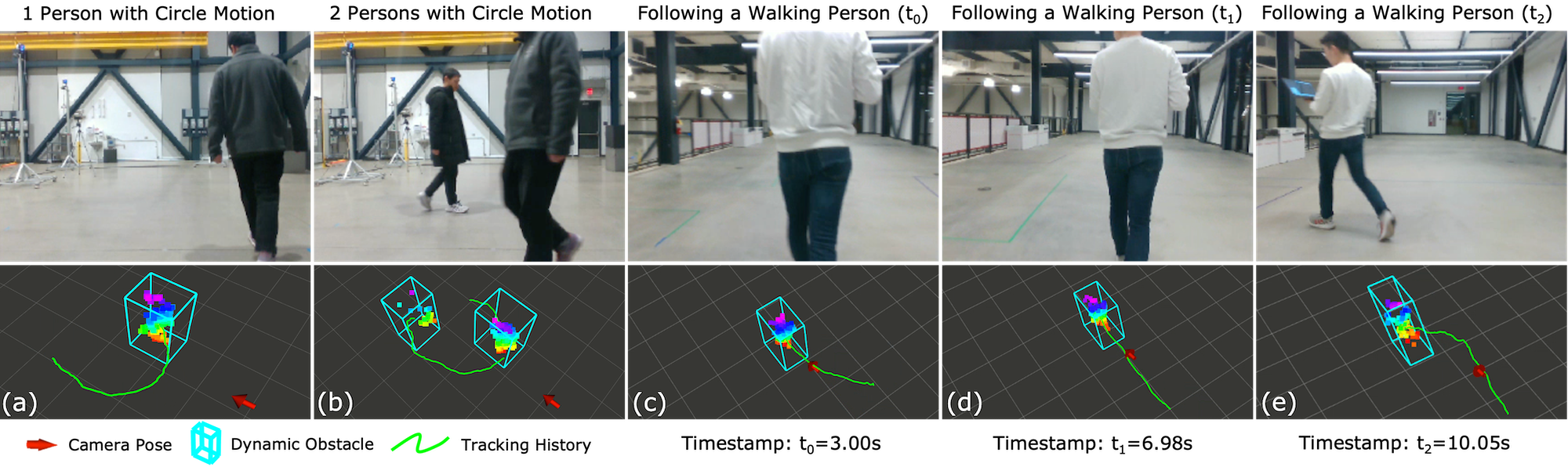}
    \caption{The dynamic obstacle detection and tracking experiments with a handheld robot camera. The blue bounding boxes containing point clouds visualize the dynamic obstacles' detection results, and the tracking histories are shown as green curves. Figures (a) and (b) show the detection results of persons walking in circles. Figures (c), (d), and (e) show the long-distance dynamic obstacle detection and tracking ability following a walking person. }
    \label{hold test}
\end{figure*}

\textbf{Handheld Experiments:} The handheld experiments are conducted by moving the robot's camera in dynamic environments to simulate the navigation trajectories. Fig. \ref{hold test} shows the example experiments with results. The first example experiment (Fig. \ref{hold test}a-b) shows persons walking in circles in front of the camera. One can see that our proposed algorithm can detect multiple persons in the camera's FOV and track the history trajectories (shown as green curves) of dynamic obstacles. Note that we only visualized the past 3 seconds' history trajectories. The second example experiment (Fig. \ref{hold test}d-e) lets the camera follow a walking person. The timestamp $\text{t}_{\text{i}}$ denotes the time starting from when the first time the dynamic obstacle is detected. The detection results show that our method can allow the robot to perform long-distance detection and tracking of the dynamic obstacle. However, from the experiment observation, we also notice that the occlusion can cause losing track of the obstacles, which is the limitation of the current system. Besides, due to the camera range limitation, the robot can only detect and track the obstacles in the camera's field of view.

\begin{figure}[t] 
    \centering
    \includegraphics[scale=1.85]{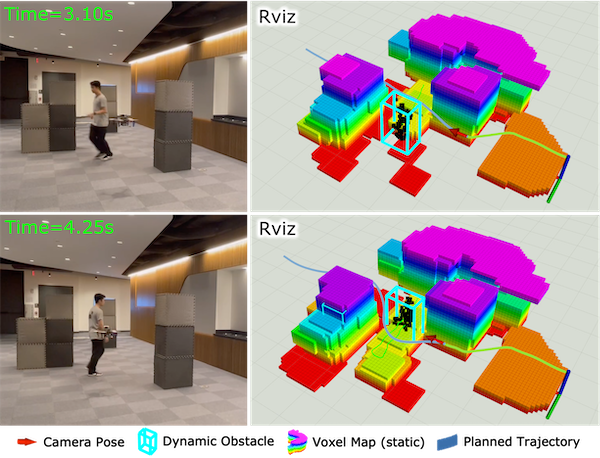}
    \caption{Autonomous robot navigation in dynamic environments using the proposed algorithm. The onboard obstacle detection results (blue bounding boxes) can help the robot modify its planned path to avoid obstacles safely.}
    \label{flight test}
\end{figure}

\textbf{Navigation Experiments:} We prepare the dynamic environment consisting of both static and dynamic obstacles to test the autonomous robot's navigation ability. The experiment is shown in Fig. \ref{flight test}. Note that the static occupancy voxel map is also used for static obstacle avoidance. In the experiment, the robot is required to navigate to the given goal position, which is 15 meters from the start location. During the navigation period, two persons (only one shown in the figure) are walking randomly as dynamic obstacles, and the robot must avoid them safely. The figure shows that the walking person is successfully detected as a dynamic obstacle, and the robot can efficiently modify its planned trajectory based on the dynamic obstacle's states.

\section{Conclusion and Future Work}
This paper presents our lightweight 3D dynamic obstacle detection and tracking (DODT) algorithm for autonomous robots navigating dynamic environments with limited computation. Our method adopts an ensemble detection strategy to obtain refined detection results by combining multiple computationally efficient but low-accuracy detectors. In addition, the proposed feature-based data association and tracking method prevents incorrect matches of obstacles with detected histories. Besides, with the obstacles' state estimations, our dynamic obstacle identification module can classify the detected obstacles into static and dynamic. Finally, we propose using the learning-based method as an optional and auxiliary module to enhance the detection range and dynamic obstacle identification. Our experimental results show that our method has the lowest position error (0.11m) and a velocity error (0.23m/s) compared with benchmarking algorithms. In the flight experiments, our method enables the robot to adapt its trajectory efficiently for dynamic collision avoidance. Future improvements can be realized through sensor fusion by harnessing the capabilities of multiple-camera systems. Furthermore, more advanced tracking techniques can be investigated to address tracking losses caused by occlusion.
 

\bibliographystyle{IEEEtran}
\bibliography{bibliography.bib}

\end{document}